\documentclass[10pt,twocolumn,letterpaper]{article}

\usepackage[pagenumbers]{iccv} % To force page numbers, e.g. for an arXiv version

\usepackage{algorithm}
\usepackage{algorithmic}
\usepackage{amssymb}
\usepackage{amsmath}
\usepackage{makecell}
\usepackage{multirow}
\usepackage{colortbl}
\usepackage{pifont}
\usepackage[dvipsnames]{xcolor}

\definecolor{mygray}{gray}{.90}
\definecolor{myblue}{RGB}{230, 245, 255}

\definecolor{iccvblue}{rgb}{0.21,0.49,0.74}
\usepackage[pagebackref,breaklinks,colorlinks,allcolors=iccvblue]{hyperref}

\def\paperID{6283} % *** Enter the Paper ID here
\def\confName{ICCV}
\def\confYear{2025}

\title{Moment Quantization for Video Temporal Grounding}

\author{Xiaolong Sun\textsuperscript{\rm 1}, Le Wang\textsuperscript{\rm 1}\thanks{Corresponding author.}\ , Sanping Zhou\textsuperscript{\rm 1} \\ Liushuai Shi\textsuperscript{\rm 1}, Kun Xia\textsuperscript{\rm 1}, Mengnan Liu\textsuperscript{\rm 1}, Yabing Wang\textsuperscript{\rm 1}, Gang Hua\textsuperscript{\rm 2}\\
\textsuperscript{\rm 1}National Key Laboratory of Human-Machine Hybrid Augmented Intelligence,\\
National Engineering Research Center for Visual Information and Applications,\\
and Institute of Artificial Intelligence and Robotics, Xi’an Jiaotong University\\
\textsuperscript{\rm 2}Multimodal Experiences Research Lab, Dolby Laboratories\\
{\tt\small \{sunxiaolong,shiliushuai,xiakun,Liumn\}@stu.xjtu.edu.cn}\\
{\tt\small \{lewang,spzhou\}@xjtu.edu.cn},
{\tt\small \{wyb7wyb7,ganghua\}@gmail.com}
}

\begin{document}
\maketitle
\begin{abstract}
Video temporal grounding is a critical video understanding task, which aims to localize moments relevant to a language description. 
The challenge of this task lies in distinguishing relevant and irrelevant moments. Previous methods focused on learning continuous features exhibit weak differentiation between foreground and background features. In this paper, we propose a novel \textbf{M}oment-\textbf{Q}uantization based \textbf{V}ideo \textbf{T}emporal \textbf{G}rounding method (MQVTG), which quantizes the input video into various discrete vectors to enhance the discrimination between relevant and irrelevant moments.
Specifically, MQVTG maintains a learnable moment codebook, where each video moment matches a codeword.
Considering the visual diversity, i.e., various visual expressions for the same moment, MQVTG treats moment-codeword matching as a clustering process without using discrete vectors, avoiding the loss of useful information from direct hard quantization. 
Additionally, we employ effective prior-initialization and joint-projection strategies to enhance the maintained moment codebook. With its simple implementation, the proposed method can be integrated into existing temporal grounding models as a plug-and-play component. Extensive experiments on six popular benchmarks demonstrate the effectiveness and generalizability of MQVTG, significantly outperforming state-of-the-art methods. Further qualitative analysis shows that our method effectively groups relevant features and separates irrelevant ones, aligning with our goal of enhancing discrimination.
\end{abstract}    
\section{Introduction}
\label{sec:intro}

The rapid expansion of short-form video content on social media platforms has made video a highly engaging multi-medium format~\cite{apostolidis2021video,mao2020generation}. The significant surge prompts users to selectively engage with brief moments of interest rather than passively watch entire videos. This derives the topic of video temporal grounding (VTG), which aims to ground video clips based on natural language descriptions. Aside from the moment retrieval~\cite{gao2017tall,lei2021detecting,liu2022umt}, recent works in VTG have also been interested in highlight detection~\cite{lin2023univtg,xiong2019less}. As a multi-modal video understanding task, variable and open-vocabulary language descriptions complicate video representation learning in VTG. This makes it difficult to distinguish between relevant and irrelevant moments, posing significant challenges to the task. %video summarization~\cite{apostolidis2021video,jiang2022joint} and

\begin{figure*}[t]
    \centering
    \includegraphics[height=4.4cm]{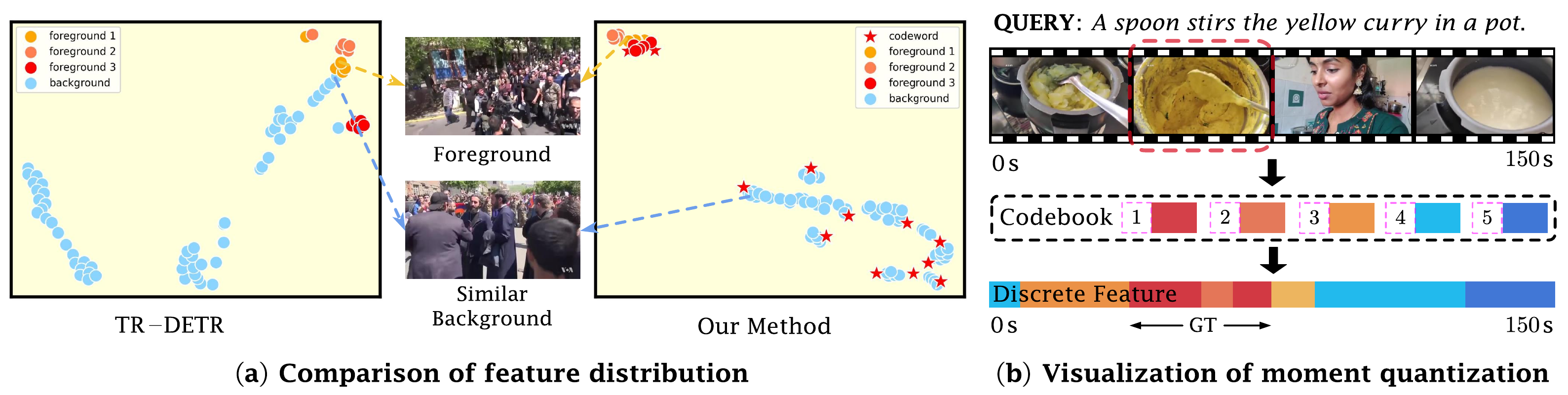}
    % \vspace{-0.4cm}
    \caption{(a) Comparison of codebook vectors, foreground and background features between TR-DETR~\cite{sun2024tr} and our method for a given example. Compared to previous methods focused on learning continuous features, our method, aided by codebook vectors, achieves better foreground aggregation and foreground-background separation. (b) Visualization of moment quantization. The moment quantization discriminates foreground and background moments. 
    The foregrounds are represented by red-related vectors, while the backgrounds are represented by other discrete vectors. 
    The quantized features bring discriminative information to generate more accurate localization. 
    % generates more accurate location results, benefiting from the moment quantization 
    % benefit from the moment quantization to  
    % thus we use continuous features with discriminative information to assist in video temporal grounding.
    }
    \label{fig:1}
    \vspace{-0.4cm}
\end{figure*}

To address this challenge, previous works~\cite{xiao2024bridging,yang2024task,jiang2024prior} learn continuous features with complex modal interactions. However, videos contain much more redundant information, while VTG aims to separate video clips into foreground and background, focusing on essential discriminative information. As shown in Fig.~\ref{fig:1}(a), we present an example with multiple non-contiguous foreground moments. Due to video information redundancy, previous methods (\eg, TR-DETR~\cite{sun2024tr}) struggle with distinguishing foreground and similar background features (\ie, red and blue points are close in feature space) and poorly grouping foreground features (\ie, different red points are far apart in feature space). Additionally, the interested foreground can be described concisely by a discrete language query~\cite{vinyals2015show,venugopalan2015sequence}.  For example in Fig.~\ref{fig:1}(b), we can accurately describe the video moment of a spoon stirring curry using the discrete language query. That motivates us to ask: \emph{Can we describe the continuous video moments via discrete vectors to enhance discrimination between relevant and irrelevant moments?} Existing works~\cite{esser2021taming,zhang2024codebook,yu2021vector} have explored various patch-level quantization on images, however, moment-level quantization on videos remains an unexplored area.

To achieve our goal, we propose a \textbf{M}oment-\textbf{Q}uantization based \textbf{V}ideo \textbf{T}emporal \textbf{G}rounding method (MQVTG), which quantizes the video moments into discrete vectors.
The foreground and background are distinguished by different vectors, thus improving the discrimination between relevant and irrelevant moments.
As shown in Fig.~\ref{fig:1}(b), the foreground moments are represented by red-related vectors, while the background moments are represented by other discrete vectors.
To construct this moment quantization, a simple implementation, named clip quantization, is proposed to quantize individual video clips into a discrete codebook similar to traditional patch quantization on images.
However, clip quantization overlooks the characteristics of the video moments: 1) A moment crosses multiple video clips (cross-clip); 2) A moment described by the same language may express various visual forms (visual diversity).

Drive by the two characteristics, the moment quantization is built on the clip quantization with three innovations.
First,  we maintain the moment codebook based on the video clips after temporal modeling rather than isolated video clips to meet the cross-clip nature of the moment.
Second,  due to the visual diversity of moments, directly using discrete codewords loses distinctive information. For example in Fig.~\ref{fig:1}(a), directly replacing all video features with limited codebook vectors is clearly inappropriate. Thus, we deliver the quantized continuous features into the downstream localization module, to retain the visual diversity.
Furthermore,  two effective prior initialization and joint projection strategies are proposed to enhance the moment codebook. 

We conduct extensive experiments on six popular video temporal grounding benchmarks to validate the effectiveness of our method, which achieves state-of-the-art performances for all benchmarks. The moment quantization also serves as a plug-and-play component, performing well in both encoder-only and encoder-decoder architectures. As shown in Fig.~\ref{fig:1}(a), our method can effectively group foreground and separate fore/background features, aligning with our goal of enhancing discrimination. Our main contributions are summarized as follows:
\begin{itemize}
    \item To our knowledge, this is the first introduction of vector quantization to the VTG task. We propose a Moment-Quantization based Video Temporal Grounding method that describes the continuous video moments via discrete vectors to enhance discrimination between moments.
    %treats vector quantization as codebook-driven feature clustering process to enhance moment discrimination.
    % We propose a Moment-Quantization based Video Temporal Grounding method, which treats moment quantization as a codebook-driven feature clustering process to enhance discrimination between moments.
    \item To adapt vector quantization from images to videos, we introduce two progressive implementations, clip quantization and moment quantization, both capable of quantizing video features to capture discriminative information.
    \item Clip quantization, as a naive implementation, simply aligns with image quantization, while moment quantization considers the characteristics of video moments, \ie, cross-clip and visual diversity.
    % \item \textcolor{red}{We improve the classic codebook module, \textit{i.e.}, initializing codebook vectors by a pre-trained vision encoder and reparameterizing these vectors via a linear projector, to adapt to moment quantization.}
    \item Extensive experiments demonstrate the effectiveness of our method, and it can be integrated into existing grounding models, showing strong generalizability.
\end{itemize}

% (1) To the best of our knowledge, it is the first time that we propose to introduce vector quantization into video temporal grounding task for discriminative video representation learning. (2) Different from previous patch quantization with discrete features, we treat moment quantization as a codebook-driven deep feature clustering process and optimize the initialization protocol and computational strategy of the codebook vector. (3) Extensive experimental results demonstrate the effectiveness of the moment quantization strategy, and it can be integrated into existing grounding models as a plug-and-play component, demonstrating good generalizability.

\section{Related Work}
\label{related}

\begin{figure*}[!ht]
    \centering
    \includegraphics[height=5.5cm]{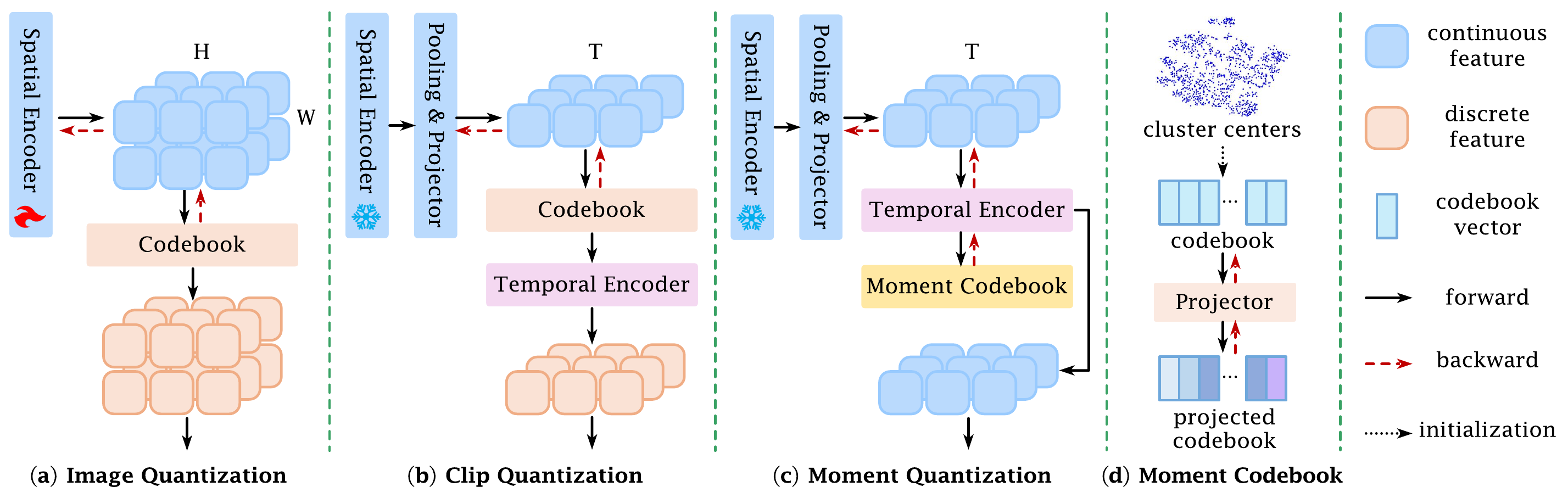}
    \caption{Comparison of three quantization methods including (a) the classic image quantization, (b) clip quantization that is a simple implementation of vector quantization for videos, and (c) the improved moment quantization for video temporal grounding. The design of the moment codebook used in moment quantization is shown in (d).}
    \label{fig:2}
    \vspace{-0.2cm}
\end{figure*}

\paragraph{Video Temporal Grounding.} Video temporal grounding can be divided into moment retrieval (MR)~\cite{gao2017tall,jang2023knowing,lei2021detecting,moon2023query} and highlight detection (HD)~\cite{badamdorj2021joint,hong2020mini,sun2014ranking}, which localizes the relevant moments and scores the clip-wise correspondence to the query. Recent research~\cite{lei2021detecting} constructs the QVHighlights dataset to facilitate joint learning of MR\&HD and proposes a baseline model based on DETR. QD-DETR~\cite{moon2023query} exploits the textual information by involving video-text pair negative relationship learning. $\text{R}^2$-Tuning~\cite{liu2024r} learns a lightweight side adapter to adaptively pool spatial details and refine temporal correlations. However, all previous methods focus on learning continuous features, overlooking the inherently discrete semantic nature of video moments. In contrast, we introduce the discrete learning approach of vector quantization to aid the VTG task.
% LLMEPET~\cite{jiang2024prior} presents an approach for enhancing video temporal grounding by integrating large language model encoders and pseudo-event regulation. Building upon this, UMT~\cite{liu2022umt} incorporates audio modality into the model, catering to scenarios for missing queries. Conventionally, moment retrieval and highlight detection are addressed in isolation.
\vspace{-0.4cm}
\paragraph{Vector Quantization.}
Vector quantization aims to represent the data with entries of a learnable codebook, which achieves discrete and compressed representation. Many researchers~\cite{mao2021discrete,csurka2004visual,nasrabadi1988image} have shown that learning discrete representation contributes to visual understanding. Recently, VQVAE~\cite{oord2018representation} uses a codebook to learn a discrete visual representation of images, which can learn the discrete feature distribution of an image effectively and is widely used in many generative models. Along this line of research, discrete representation learning has been widely used in many vision tasks~\cite{esser2021taming,gu2022vector,yang2023vector,zhang2024codebook}. Existing works have explored various quantization methods on images or audio, however, vector quantization for videos remains an unexplored area. % For example, Yang \textit{et al.}~\cite{yang2023vector} presents a plug-and-play method for low-quality image recognition through vector quantization and self-attention. QDFormer~\cite{li2024qdformer} proposes a semantic decomposition method based on product quantization to provide a more robust audio representation for audiovisual interaction in complex environments.

\section{Method}
\label{method}

Given an untrimmed video $\mathcal{V}$ with $T$ frames and an associated natural language description $ \mathcal{Q}$ with $N$ words, video temporal grounding aims to predict clip-wise saliency scores and localize the center coordinate and span of target moments that are most relevant to the description. Our goal is to introduce the discrete learning approach in vector quantization to video temporal grounding, aiming to enhance discrimination between relevant and irrelevant moments. Thus we propose a moment quantization method for video temporal grounding, called MQVTG. 

In this section, we first introduce the preliminaries of image quantization in Sec.~\ref{3.1}. Then, we propose the clip quantization in Sec.~\ref{3.2}, which is a simple implementation of vector quantization for videos. The improved moment quantization and its moment codebook are explained in detail in Sec.~\ref{3.3} and Sec.~\ref{3.4}. Finally, the overall architecture and training objectives of our method are introduced in Sec.~\ref{3.5} and Sec.~\ref{3.6}, respectively.

\subsection{Preliminaries: Image Quantization} \label{3.1}
An image quantization model is typically a reconstructive encoder-decoder architecture that includes a codebook module to convert continuous representations into discrete ones as illustrated in Fig.~\ref{fig:2} (a). Formally, the spatial encoder maps the input image into a latent space, producing a continuous feature $z \in \mathbb{R}^{H\times W\times d}$, where $d$ denotes the hidden dimension. The feature is then quantized using a learnable codebook $C \in \mathbb{R}^{K\times d}$, where $K$ represents the number of codewords in the codebook. The codebook module selects the nearest codeword by minimizing the distance between $z$ and the codewords:
\begin{equation}
    \hat{z} = C(z) = c_k, \;\text{where} \ k=\text{arg}\mathop{\text{min}}\limits_{i} \left | \left | z - c_i \right | \right |^2_2,
    \label{minimize}
\end{equation}
where $c_i$ is the i-th codeword. $c_k$ is usually denoted by $\hat{z}$, that is, the quantized discrete feature passed to the decoder to reconstruct the input image. The most classic image quantization model is VQ-VAE~\cite{van2017neural}, which uses a straight-through estimator~\cite{bengio2013estimating} with codebook loss and commitment loss to narrow the gap between the selected codewords and the encoder output. Image quantization has been widely studied~\cite{esser2021taming,zhang2024codebook,yu2021vector}, however, vector quantization on videos remains an unexplored area.

\subsection{Clip Quantization} \label{3.2}
From the above perspective, the codebook update of the vector quantization model is intrinsically a dictionary learning process. In other words, the codebook training is like $k$-means clustering, where cluster centers are the discrete codewords. The feature discrimination formed by the clustering process aligns with the requirement of the VTG task to distinguish between relevant and irrelevant moments. Based on this, we propose two progressive implementations of vector quantization for videos, clip and moment quantization. Both of them quantize video features to capture discriminative information, while clip quantization simply aligns with image quantization and moment quantization considers the cross-clip nature and visual diversity of video moments. We first introduce clip quantization in Fig.~\ref{fig:2} (b).
% Based on this motivation, we present clip quantization as illustrated in Fig.~\ref{fig:2} (b), which is a simple implementation of vector quantization for videos. To adapt vector quantization from images to videos, we focus on three main considerations: 1) where to apply quantization, 2) how to leverage discrete and continuous features, and 3) how codebook requirements for videos differ from images. As a simple implementation, clip quantization aligns with image quantization in these three aspects.

\vspace{-0.4cm}
\paragraph{Quantization before Temporal Modeling.} Unlike previous image quantization, which focuses on spatial patches, we aim to perform quantization on temporal video clips. As shown in Fig.~\ref{fig:2} (b), after passing through a frozen spatial encoder, a pooling layer and a linear projector, we obtain the pooled visual feature $z_s \in \mathbb{R}^{T\times d}$, where $T$ is the number of video clips. Subsequently, a codebook module is used to quantize the continuous visual feature $z_s$ by Eq.~\ref{minimize}, which is supervised by the codebook loss $\mathcal{L}_{\text{cb}}$ and the commitment loss $\mathcal{L}_{\text{cmt}}$~\cite{van2017neural}. The quantized discrete feature is then sent to a multi-modal temporal encoder $E_{t}$ to generate the video feature for temporal grounding. For simplicity, we omit the text input and multi-modal interaction process.

The difference between clip and image quantization is that the former quantizes individual video clips, while the latter quantizes spatial image patches. By optimizing the projector after the frozen spatial encoder with the codebook and commitment loss, clip quantization enables feature clustering for these individual video clips. However, as a simple implementation of vector quantization for videos, it lacks careful consideration for video moments. Therefore, we propose a more comprehensive quantization method for videos in the next section, called moment quantization. %\textcolor{red}{to enhance the discrimination between relevant and irrelevant moments.}

\subsection{Moment Quantization} \label{3.3}
The implementation of our moment quantization is illustrated in Fig.~\ref{fig:2} (c). Compared to the simple clip quantization, it introduces three main improvements: 1) We choose to apply quantization after temporal modeling to meet the cross-clip nature of moments; 2) We introduce a soft quantization operation instead of directly using discrete features to accommodate visual diversity; 3) We propose a moment codebook to better adapt moment quantization. 
In this section, we first explain the first two points, and the moment codebook module will be discussed in Sec.~\ref{3.4}.
% As discussed in the introduction, previous methods~\cite{moon2023query,xiao2024bridging,liu2024r} directly use the continuous video representation $z$ for prediction and localization, overlooking the discrete nature of moments, which leads to lower discrimination between different moments. Inspired by vector quantization, we introduce a carefully designed codebook module after the multimodal encoder $E$ to quantize the continuous video representation $z$. Details of the codebook improvements will be discussed in Sec.~\ref{3.3}. % For image quantization, each codeword represents a patch prototype to reconstruct the image. we consider each codeword in moment quantization as a video clip prototype, with the feature-codeword matching guided by Eq.~\ref{minimize} treated as a clustering process.
\vspace{-0.4cm}
\paragraph{Quantization after Temporal Modeling.} The clip quantization quantizes video clips before the temporal encoder $E_t$. Without temporal modeling, these video clips remain isolated and fail to form a complete moment—representing a full action or event, which conflicts with the cross-clip nature of moments. Therefore, moment quantization is applied after temporal modeling. As shown in Fig.~\ref{fig:2} (c), the pooled visual feature $z_s$ is first sent to the temporal encoder $E_{t}$ to generate the semantic-aware continuous video feature $z_t = E_t(z_s)$. Subsequently, we use the moment codebook module (discussed in Sec.~\ref{3.4}) to quantize the continuous feature $z_t$. The process is supervised by the codebook loss $\mathcal{L}_{\text{cb}}$ and the commitment loss $\mathcal{L}_{\text{cmt}}$~\cite{van2017neural}. $\mathcal{L}_{\text{cb}}$ for updating the codebook parameters is as follows:
\begin{equation}
    \mathcal{L}_{\text{cb}} = \left | \left | \hat{z_t} - \text{sg}(z_t) \right |  \right |^2_2 = \left | \left | C(z_t) - \text{sg}(E_t(z_s)) \right |  \right |^2_2,
    \label{cb}
\end{equation}
where $\text{sg}(\cdot)$ represents the stop-gradient operation and $\hat{z_t}$ is the quantized discrete vector from the codebook. Similarly, $\mathcal{L}_{\text{cmt}}$ can be formulated as:
\begin{equation}
    \mathcal{L}_{\text{cmt}} = \left | \left | \text{sg}(\hat{z_t}) - z_t \right |  \right |^2_2 = \left | \left | \text{sg}(C(z_t)) - E_t(z_s) \right |  \right |^2_2,
    \label{cmt}
\end{equation}
which is used to optimize the parameters in temporal encoder $E_t$ and make the output of $E_t$ consistent with the codebook embedding space. The two losses jointly guide the feature-codeword clustering process.

\vspace{-0.4cm}
\paragraph{Soft Quantization with Continuous Features.} Image quantization usually replaces the continuous feature with the quantized discrete vector directly and inputs it into the subsequent decoder. However, videos have greater visual diversity and information density compared to images, which means that the visual presentation of the same language description can vary greatly in video format. In this case, the discrete vectors from a limited-capacity codebook cannot accurately represent an entire video as they do for an image. On the other hand, due to the direct replacement operation in Eq.~\ref{minimize} and the poorly optimized codebook in the early training stage, some useful information may inevitably be lost~\cite{yang2023vector}, thus affecting the final localization. We will explain this in detail in Sec.~\ref{4.6}. Based on these considerations, we introduce a soft quantization operation. As shown in Fig.~\ref{fig:2} (c), we continue using the continuous features $z_t$ instead of the quantized discrete vector $\hat{z_t}$ for subsequent localization. We argue that the feature-codeword clustering process, driven by Eq.~\ref{cb} and Eq.~\ref{cmt}, enables the continuous feature $z_t$ to learn discriminative information, whereas directly using discrete vectors may actually be detrimental. The designed ablation experiments in Sec.~\ref{4.5} support our perspective.

% \vspace{-0.4cm}
% \paragraph{\textcolor{red}{Comparison with Other Quantization Methods.}} We compare our moment quantization with other two methods in Fig.~\ref{fig:2}. Previous image quantization (\textit{e.g.},~\cite{van2017neural,yu2021vector,zhang2024codebook,esser2021taming}, Fig.~\ref{fig:2}a) optimize the spatial encoder and apply direct hard quantization to generate discretized spatial patch features for image reconstruction. In contrast, our moment quantization focuses on video features after temporal modeling, with the spatial encoder frozen and only the temporal encoder optimized. The soft quantization operation would not lose useful information while learning discriminative information. \textcolor{red}{Another method, called clip quantization (Fig.~\ref{fig:2}b), quantizes video features before the temporal encoder, with codebook loss optimizing only the linear projector after the frozen spatial encoder. As video clips are independent before temporal modeling, this approach conflicts with our concept of moment quantization. We empirically compare these three methods in Sec.~\ref{experiments}.}

\begin{table*}[ht]\small
\tabcolsep=0.063cm
\begin{center}
\renewcommand{\arraystretch}{1.0}
%\resizebox{\textwidth}{!}{
\begin{tabular}{lcccc>{\columncolor{myblue}}ccccccc>{\columncolor{myblue}}ccc}
\toprule
\multirow{3}{*}{\makecell{ \\ \textbf{Method}}} & \multicolumn{7}{c}{\textbf{QVHighlights-val}} & \multicolumn{7}{c}{\textbf{QVHighlights-test}}\\
\cmidrule(rl){2-8} \cmidrule(rl){9-15}
& \multicolumn{2}{c}{MR-R$\mathrm{1}$} & \multicolumn{3}{c}{MR-mAP} & \multicolumn{2}{c}{HD} & \multicolumn{2}{c}{MR-R$\mathrm{1}$} & \multicolumn{3}{c}{MR-mAP} & \multicolumn{2}{c}{HD}\\
\cmidrule(rl){2-3} \cmidrule(rl){4-6} \cmidrule(rl){7-8} \cmidrule(rl){9-10} \cmidrule(rl){11-13} \cmidrule(rl){14-15}
& @$\mathrm{0.5}$ & @$\mathrm{0.7}$ & @$\mathrm{0.5}$ & @$\mathrm{0.75}$ & Avg. & mAP & HIT@$\mathrm{1}$ & @$\mathrm{0.5}$ & @$\mathrm{0.7}$ & @$\mathrm{0.5}$ & @$\mathrm{0.75}$ & Avg. & mAP & HIT@$\mathrm{1}$ \\
\addlinespace[1pt]
% \noalign{\smallskip}
\hline
% \noalign{\smallskip}
\addlinespace[1pt]
M-DETR~\cite{lei2021detecting}{\scriptsize \textit{NeurIPS'21}} & $\mathrm{53.94}$ & $\mathrm{34.84}$ & $\mathrm{-}$ & $\mathrm{-}$ & $\mathrm{32.20}$ & $\mathrm{35.65}$ & $\mathrm{55.55}$ & $\mathrm{52.89}$ & $\mathrm{33.02}$ & $\mathrm{54.82}$ & $\mathrm{29.40}$ & $\mathrm{30.73}$ & $\mathrm{35.69}$ & $\mathrm{55.60}$\\

UMT~\cite{liu2022umt}{\scriptsize \textit{CVPR'22}} & $\mathrm{60.26}$ & $\mathrm{44.26}$ & $\mathrm{-}$ & $\mathrm{-}$ & $\mathrm{38.59}$ & $\mathrm{39.85}$ & $\mathrm{64.19}$ & $\mathrm{56.23}$ & $\mathrm{41.18}$ & $\mathrm{53.83}$ & $\mathrm{37.01}$ & $\mathrm{36.12}$ & $\mathrm{38.18}$ & $\mathrm{59.99}$\\

QD-DETR~\cite{moon2023query}{\scriptsize \textit{CVPR'23}} & $\mathrm{62.68}$ & $\mathrm{46.66}$ & $\mathrm{62.23}$ & $\mathrm{41.82}$ & $\mathrm{41.22}$ & $\mathrm{39.13}$ & $\mathrm{63.03}$ & $\mathrm{62.40}$ & $\mathrm{44.98}$ & $\mathrm{62.52}$ & $\mathrm{39.88}$ & $\mathrm{39.86}$ & $\mathrm{38.94}$ & $\mathrm{62.40}$ \\

UniVTG~\cite{lin2023univtg}{\scriptsize \textit{ICCV'23}} & $\mathrm{59.74}$ & $\mathrm{-}$ & $\mathrm{-}$ & $\mathrm{-}$ & $\mathrm{36.13}$ & $\mathrm{38.83}$ & $\mathrm{61.81}$ & $\mathrm{58.86}$ & $\mathrm{40.86}$ & $\mathrm{57.60}$ &  $\mathrm{35.59}$ & $\mathrm{35.47}$ & $\mathrm{38.20}$ & $\mathrm{60.96}$\\

% EaTR~\cite{jang2023knowing}{\scriptsize \textit{ICCV'23}} & $\mathrm{61.36}$ & $\mathrm{45.79}$ & $\mathrm{61.86}$ & $\mathrm{41.91}$ & $\mathrm{41.74}$ & $\mathrm{37.15}$ & $\mathrm{58.65}$ & $\mathrm{-}$ & $\mathrm{-}$ & $\mathrm{-}$ & $\mathrm{-}$ & $\mathrm{-}$ & $\mathrm{-}$ & $\mathrm{-}$\\

% MomentDiff~\cite{li2024momentdiff}{\scriptsize \textit{NeurIPS'23}}  & $\mathrm{-}$ & $\mathrm{-}$ & $\mathrm{-}$ & $\mathrm{-}$ & $\mathrm{-}$ & $\mathrm{-}$ & $\mathrm{-}$ & $\mathrm{57.42}$ & $\mathrm{39.66}$ & $\mathrm{54.02}$ & $\mathrm{35.73}$ & $\mathrm{35.95}$ & $\mathrm{-}$ & $\mathrm{-}$\\

% MESM~\cite{liu2024towards}{\scriptsize \textit{AAAI'24}} & - & - & - & - & - & - & - & 62.78 & 45.20 & 62.64 & 41.45 & 40.68 & - & -\\

TR-DETR~\cite{sun2024tr}{\scriptsize \textit{AAAI'24}} & $\mathrm{67.10}$ & $\mathrm{51.48}$ & $\mathrm{66.27}$ & $\mathrm{46.42}$ & $\mathrm{45.09}$ & $\mathrm{40.55}$ & $\mathrm{64.77}$ & $\mathrm{64.66}$ & $\mathrm{48.96}$ & $\mathrm{63.98}$ & $\mathrm{43.73}$ & $\mathrm{42.62}$ & $\mathrm{39.91}$ & $\mathrm{63.42}$\\

CG-DETR~\cite{moon2023correlation}{\scriptsize \textit{Arxiv'24}} & $\mathrm{67.35}$ & $\mathrm{52.06}$ & $\mathrm{65.57}$ & $\mathrm{45.73}$ & $\mathrm{44.93}$ & $\mathrm{\boldsymbol{40.79}}$ & $\mathrm{\boldsymbol{66.71}}$ & $\mathrm{65.43}$ & $\mathrm{48.38}$ & $\mathrm{64.51}$ & $\mathrm{42.77}$ & $\mathrm{42.86}$ & $\mathrm{40.33}$ & $\mathrm{\boldsymbol{66.21}}$ \\

UVCOM~\cite{xiao2024bridging}{\scriptsize \textit{CVPR'24}} & $\mathrm{65.10}$ & $\mathrm{51.81}$ & $\mathrm{-}$ & $\mathrm{-}$ & $\mathrm{45.79}$ & $\mathrm{40.03}$ & $\mathrm{63.29}$ & $\mathrm{63.55}$ & $\mathrm{47.47}$ & $\mathrm{63.37}$ & $\mathrm{42.67}$ & $\mathrm{43.18}$ & $\mathrm{39.74}$ & $\mathrm{64.20}$\\

TaskWeave~\cite{yang2024task}{\scriptsize \textit{CVPR'24}} & $\mathrm{64.26}$ & $\mathrm{50.06}$ & $\mathrm{65.39}$ & $\mathrm{46.47}$ & $\mathrm{45.38}$ & $\mathrm{39.28}$ & $\mathrm{63.68}$ & $\mathrm{-}$ & $\mathrm{-}$ & $\mathrm{-}$ & $\mathrm{-}$ & $\mathrm{-}$ & $\mathrm{-}$ & $\mathrm{-}$\\

$\text{R}^2$-Tuning~\cite{liu2024r}{\scriptsize \textit{ECCV'24}} & $\mathrm{\boldsymbol{68.71}}$ & $\mathrm{52.06}$ & $\mathrm{-}$ & $\mathrm{-}$ & $\mathrm{47.59}$ & $\mathrm{\underline{40.59}}$ & $\mathrm{64.32}$ & $\mathrm{\boldsymbol{68.03}}$ & $\mathrm{49.35}$ & $\mathrm{\boldsymbol{69.04}}$ & $\mathrm{\underline{47.56}}$ & $\mathrm{\underline{46.17}}$ & $\mathrm{\boldsymbol{40.75}}$ & $\mathrm{64.20}$ \\

BAM-DETR~\cite{lee2023bam}{\scriptsize \textit{ECCV'24}} & $\mathrm{65.10}$ & $\mathrm{51.61}$ & $\mathrm{65.41}$ & $\mathrm{\underline{48.56}}$ & $\mathrm{\underline{47.61}}$ & $\mathrm{-}$ & $\mathrm{-}$ & $\mathrm{62.71}$ & $\mathrm{48.64}$ & $\mathrm{64.57}$ & $\mathrm{46.33}$ & $\mathrm{45.36}$ & $\mathrm{-}$ & $\mathrm{-}$ \\

LLMEPET~\cite{jiang2024prior}{\scriptsize \textit{ACMMM'24}} & $\mathrm{66.58}$ & $\mathrm{51.10}$ & $\mathrm{-}$ & $\mathrm{-}$ & $\mathrm{46.24}$ & $\mathrm{-}$ & $\mathrm{-}$ & $\mathrm{\underline{66.73}}$ & $\mathrm{\underline{49.94}}$ & $\mathrm{65.76}$ & $\mathrm{43.91}$ & $\mathrm{44.05}$ & $\mathrm{\underline{40.33}}$ & $\mathrm{\underline{65.69}}$\\

SpikeMba~\cite{li2024spikemba}{\scriptsize \textit{Arxiv'24}} & $\mathrm{65.32}$ & $\mathrm{51.33}$ & $\mathrm{-}$ & $\mathrm{44.96}$ & $\mathrm{44.84}$ & $\mathrm{-}$ & $\mathrm{-}$ & $\mathrm{64.13}$ & $\mathrm{49.42}$ & $\mathrm{-}$ & $\mathrm{43.67}$ & $\mathrm{43.79}$ & $\mathrm{-}$ & $\mathrm{-}$ \\

% \textbf{RGTR (n = 10)} & \textbf{66.08} & \underline{48.64} & \underline{65.78} & 42.89 & 43.09 &\underline{67.61} & \underline{52.19} & \underline{67.07} & 45.86 & 45.59 \\
RGTR~\cite{sun2024diversifying}{\scriptsize \textit{AAAI'25}} & $\mathrm{67.68}$ & $\mathrm{\underline{52.90}}$ & $\mathrm{\underline{67.38}}$ & $\mathrm{48.00}$ & $\mathrm{46.95}$ & $\mathrm{-}$ & $\mathrm{-}$ & $\mathrm{65.50}$ & $\mathrm{49.22}$ & $\mathrm{67.12}$ & $\mathrm{45.77}$ & $\mathrm{45.53}$ & $\mathrm{-}$ & $\mathrm{-}$\\

% \textbf{Quantize} & \textbf{68.77} & \textbf{52.32} & \textbf{68.88} & \textbf{51.32} & \textbf{48.76} & 66.67 & 48.90 & \underline{67.70} & \textbf{48.68} & \textbf{46.89}\\
\rowcolor{myblue}
% Quantize(0.2) & 67.55 & 52.32 & 68.49 & 50.64 & 48.44 & - & - & - & - & - \\
% Quantize(0.2 + 0.5) & 69.10 & 51.55 & 69.12 & 50.10 & 48.32 & - & - & - & - & - \\
% Quantize(0.2 + 0.5) & 68.84 & 51.74 & 69.21 & 50.34 & 48.38 & - & - & - & - & - \\
\textbf{MQVTG (Ours)} & $\mathrm{\underline{67.94}}$ & $\mathrm{\boldsymbol{53.03}}$ & $\mathrm{\boldsymbol{68.54}}$ & $\mathrm{\boldsymbol{51.48}}$ & $\mathrm{\boldsymbol{48.81}}$ & $\mathrm{40.23}$ & $\mathrm{\underline{65.29}}$ & $\mathrm{66.28}$ & $\mathrm{\boldsymbol{50.00}}$ & $\mathrm{\underline{67.98}}$ & $\mathrm{\boldsymbol{48.55}}$ & $\mathrm{\boldsymbol{47.08}}$ & $\mathrm{39.49}$ & $\mathrm{64.07}$ \\
% \rowcolor{myblue}
% Quantize & \underline{67.81} & 52.00 & \textbf{68.87} & \textbf{51.44} & \textbf{48.94} & 40.25 & 64.45 & 66.47 & 49.55 & 68.44 & 49.59 & 47.29 & 39.48 & 63.81 \\
% \rowcolor{myblue}
% Quantize & \underline{67.81} & 52.00 & \textbf{68.87} & \textbf{51.44} & \textbf{48.94} & 40.25 & 64.45 & 66.47 & 49.55 & 69.05 & 49.13 & 47.01 & 39.48 & 63.81 \\
%\rowcolor{myblue}
%\textbf{Quantize 2} & 66.08 & 47.92 & \underline{67.43} & \textbf{48.51} & \textbf{46.22} & \textbf{69.10} & \textbf{52.97} & \textbf{69.04} & \textbf{51.05} & \textbf{48.88}\\
\toprule
\end{tabular}
%}
\caption{Moment retrieval (MR) and highlight detection (HD) results on QVHighlights \textit{val} and \textit{test} splits.} %with the features from Slowfast and CLIP.} %We highlight the best score in each column in \textbf{bold}, and the second best score with \underline{underline}.}
\label{table1}
\vspace{-0.4cm}
\end{center}
\end{table*}

\begin{figure}[t]
    \centering
    \includegraphics[width=8.4cm]{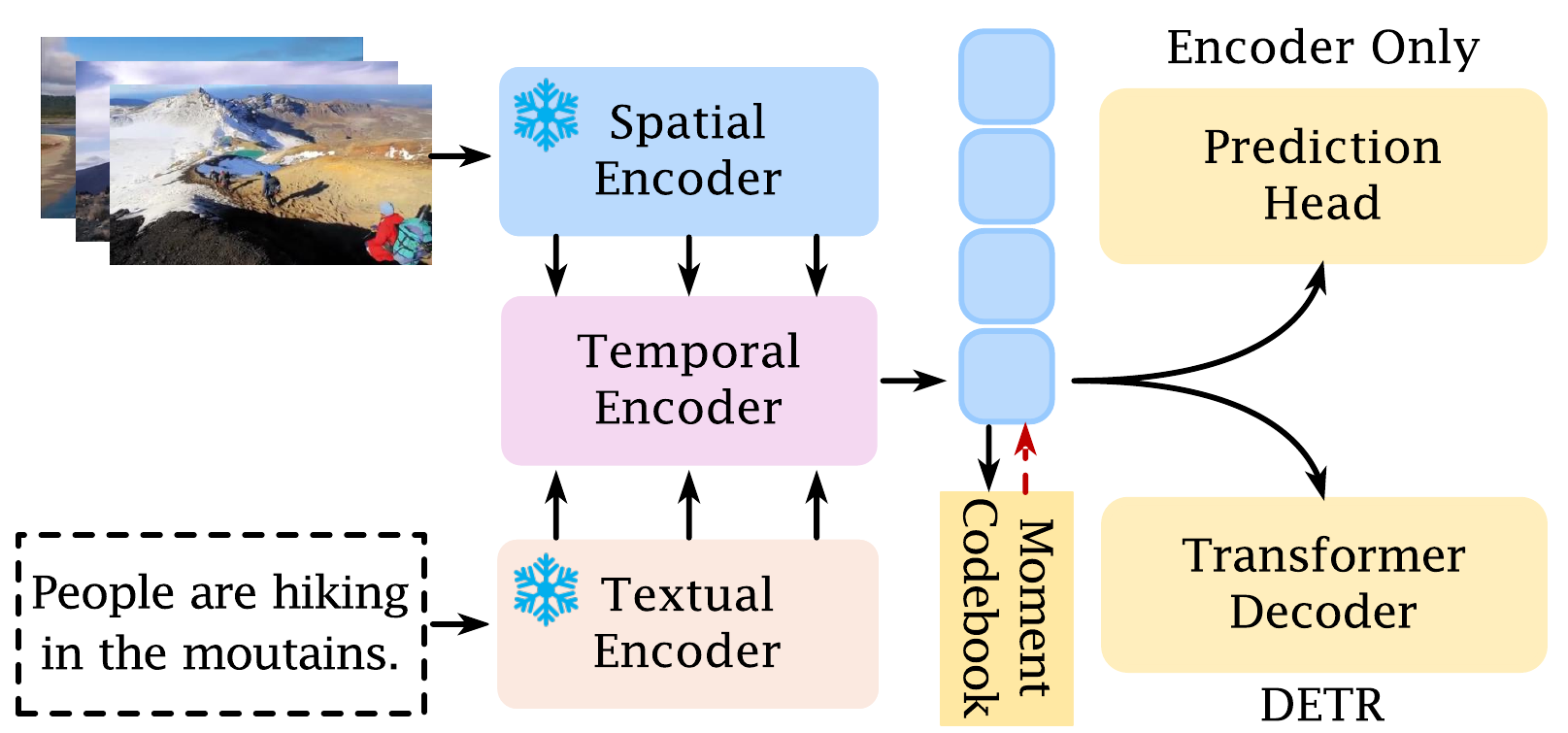}
    \caption{The architectures of MQVTG, including the encoder-only architecture and encoder-decoder (DETR) architecture.}
    \label{fig:3}
    \vspace{-0.1cm}
\end{figure}

\subsection{Moment Codebook} \label{3.4}

In the previous image quantization models~\cite{van2017neural,esser2021taming,yu2021vector}, the codebook vectors are initialized randomly. As shown in Eq.~\ref{minimize}, during each iteration, only a small subset of codebook vectors related to the current training batch are optimized. Therefore, the random initialization causes only a few frequently optimized codebook vectors to align with the feature distribution generated by the temporal encoder. This conflicts with the rich semantic representation needed for video moments. On the other hand, different codebook vectors are independent of each other in previous methods. However, the video clips from the same moment are semantically related. We establish this temporal semantic correlation via the temporal encoder and believe it should extend to the codebook vectors as well. Based on the above discussion, we propose a moment codebook, as illustrated in Fig.~\ref{fig:2} (d), which employs effective prior-initialization and joint-projection strategies to adapt moment quantization.

\vspace{-0.5cm}
\paragraph{Prior Initialization.} We first extract patch-level features for each clip in each video from the training dataset via a pre-trained visual encoder (\eg, the CLIP model), and then obtain the clip-level features by spatially max-pooling. Since the process of codebook training is basically the process of finding cluster centers, we directly employ $k$-means clustering on all clip-level features and utilize the cluster centers as priors to initialize the codebook $C$. This ensures that the codebook is composed of valid latent codes.

\vspace{-0.5cm}
\paragraph{Joint Projection.} Unlike previous methods, which optimize codebook vectors directly, our moment codebook involves training a projector $P(\cdot)$. It is implemented as a simple linear layer to explore correlations between different codebook vectors, aligning with the temporal correlations of video clip features. Specifically, we replace the original codebook $C$ with the projected codebook ${C}' = P(C)$.

\subsection{The Architecture of MQVTG} \label{3.5}
Based on the above discussion, we propose the Moment-Quantization based Video Temporal Grounding method (MQVTG) as illustrated in Fig~\ref{fig:3}, which can be divided into two architectures: encoder-only and encoder-decoder. Regardless of the architecture, our quantization method adds only minimal parameters during training (\ie, the codebook parameters) and incurs no extra cost during inference.

\vspace{-0.5cm}
\paragraph{Encoder-Only.} Following previous methods~\cite{moon2023query,liu2024r}, we use the pre-trained CLIP model as the spatial encoder and textual encoder. Then, to obtain the semantic-aware video representations $z_t$, we utilize a lightweight recurrent structure~\cite{liu2024r} as the temporal encoder, mainly consisting of cross and self-attention blocks to integrate multi-layer CLIP features. As discussed in Sec.~\ref{3.3} and Sec.~\ref{3.4}, $z_t$ is sent to the moment codebook module to learn the discriminative information, supervised by Eq.~\ref{cb} and Eq.~\ref{cmt}. Finally, following the simple design from~\cite{lin2023univtg}, we adopt three heads on continuous feature $z_t$ to separately predict the classification confidence score for each frame, the boundary displacements for start-end timestamps, and the saliency score. Without further specification, we employ the encoder-only architecture for the following experiments.

\vspace{-0.5cm}
\paragraph{Encoder-Decoder.} Since our moment quantization is easy to implement, it can be easily integrated into the general encoder-decoder (DETR) architecture. As shown in Fig.~\ref{fig:3}, replacing our temporal encoder with a transformer encoder and prediction heads with a transformer decoder forms the existing general DETR architecture for VTG~\cite{moon2023query,sun2024tr,yang2024task}. The moment codebook remains positioned between the encoder and decoder, adding discriminative information to video features. The experiment in Sec.~\ref{4.4} demonstrates the generalizability of our method.

\begin{table*}[ht]\small
\tabcolsep=0.077cm
\begin{center}
\renewcommand{\arraystretch}{1.0}
\begin{tabular}{lccc>{\columncolor{myblue}}cccc>{\columncolor{myblue}}cccc>{\columncolor{myblue}}c}
    \toprule
    \multirow{2}{*}{\textbf{Method}} & \multicolumn{4}{c}{\textbf{Charades-STA}} & \multicolumn{4}{c}{\textbf{TACoS}} & \multicolumn{4}{c}{\textbf{Ego4D-NLQ}}\\
    \cmidrule(rl){2-5} \cmidrule(rl){6-9} \cmidrule(rl){10-13}
    & R$\mathrm{1}$@$\mathrm{0.3}$ & R$\mathrm{1}$@$\mathrm{0.5}$ & R$\mathrm{1}$@$\mathrm{0.7}$ & mIoU & R$\mathrm{1}$@$\mathrm{0.3}$ & R$\mathrm{1}$@$\mathrm{0.5}$ & R$\mathrm{1}$@$\mathrm{0.7}$ & mIoU & R$\mathrm{1}$@$\mathrm{0.3}$ & R$\mathrm{1}$@$\mathrm{0.5}$ & R$\mathrm{1}$@$\mathrm{0.7}$ & mIoU\\
    % \noalign{\smallskip}
    \hline
    % \noalign{\smallskip}
    \addlinespace[1pt]
    % 2D-TAN~\cite{zhang2020learning} & 40.01 & 27.99 & 12.92 & 27.22 & 58.76 & 46.02 & 27.50 & 41.25 \\
    % M-DETR~\cite{lei2021detecting} & 65.83 & 52.07 & 30.59 & 45.54 \\
    % MomentDiff~\cite{li2024momentdiff} & - & 55.57 & 32.42 & - \\
    % QD-DETR~\cite{moon2023query} & - & 57.31 & 32.55 & - \\
    2D-TAN~\cite{zhang2020learning}{\scriptsize \textit{AAAI'20}} & $\mathrm{58.76}$ & $\mathrm{46.02}$ & $\mathrm{27.50}$ & $\mathrm{41.25}$ & $\mathrm{40.01}$ & $\mathrm{27.99}$ & $\mathrm{12.92}$ & $\mathrm{27.22}$ & $\mathrm{4.33}$ & $\mathrm{1.83}$ & $\mathrm{0.60}$ & $\mathrm{3.39}$\\
    VSLNet~\cite{zhang2020span}{\scriptsize \textit{ACL'20}} & $\mathrm{60.30}$ & $\mathrm{42.69}$ & $\mathrm{24.14}$ & $\mathrm{41.58}$ & $\mathrm{35.54}$ & $\mathrm{23.54}$ & $\mathrm{13.15}$ & $\mathrm{24.99}$ & $\mathrm{4.54}$ & $\mathrm{2.40}$ & $\mathrm{1.01}$ & $\mathrm{3.54}$\\
    M-DETR~\cite{lei2021detecting}{\scriptsize \textit{NeurIPS'21}} & $\mathrm{65.83}$ & $\mathrm{52.07}$ & $\mathrm{30.59}$ & $\mathrm{45.54}$ & $\mathrm{37.97}$ & $\mathrm{24.67}$ & $\mathrm{11.97}$ & $\mathrm{25.49}$ & $\mathrm{4.34}$ & $\mathrm{1.81}$ & $\mathrm{0.65}$ & $\mathrm{3.53}$\\
    % MomentDiff~\cite{li2024momentdiff} & - & 55.57 & 32.42 & - \\
    % QD-DETR~\cite{moon2023query} & - & 57.31 & 32.55 & - \\
    UniVTG~\cite{lin2023univtg}{\scriptsize \textit{ICCV'23}} & $\mathrm{70.81}$ & $\mathrm{58.01}$ & $\mathrm{35.65}$ & $\mathrm{50.10}$ & $\mathrm{\underline{51.44}}$ & $\mathrm{34.97}$ & $\mathrm{17.35}$ & $\mathrm{33.60}$ & $\mathrm{\boldsymbol{7.28}}$ & $\mathrm{3.95}$ & $\mathrm{1.32}$ & $\mathrm{4.91}$ \\
    
    % CG-DETR~\cite{moon2023correlation} & 70.43 & 58.44 & 36.34 & 50.13 & 52.23 & \underline{39.61} & 22.23 & 36.48 \\
    % TR-DETR~\cite{sun2024tr} & $\mathrm{-}$ & $\mathrm{57.61}$ & $\mathrm{33.52}$ & $\mathrm{-}$ & $\mathrm{-}$ & $\mathrm{-}$ & $\mathrm{-}$ & $\mathrm{-}$\\

    % UVCOM~\cite{xiao2024bridging}{\scriptsize \textit{CVPR'23}} & $\mathrm{-}$ & $\mathrm{\underline{59.25}}$ & $\mathrm{36.64}$ & $\mathrm{-}$ & $\mathrm{-}$ & $\mathrm{36.39}$ & $\mathrm{23.32}$ & $\mathrm{-}$ & $\mathrm{-}$ & $\mathrm{-}$ & $\mathrm{-}$ & $\mathrm{-}$\\
    
    % RGTR~\cite{sun2024diversifying} & 72.04 & 57.93 & 35.16 & 50.32 \\
    $\text{R}^2$-Tuning~\cite{liu2024r}{\scriptsize \textit{ECCV'24}} & $\mathrm{70.91}$ & $\mathrm{\boldsymbol{59.78}}$ & $\mathrm{\underline{37.02}}$ & $\mathrm{\underline{50.86}}$ & $\mathrm{49.71}$ & $\mathrm{\underline{38.72}}$ & $\mathrm{\underline{25.12}}$ & $\mathrm{35.92}$ & $\mathrm{\underline{7.20}}$ & $\mathrm{\underline{4.49}}$ & $\mathrm{\underline{2.12}}$ & $\mathrm{\underline{4.94}}$ \\
    
    LLMEPET~\cite{jiang2024prior}{\scriptsize \textit{ACMMM'24}} & $\mathrm{\underline{70.91}}$ & $\mathrm{-}$ & $\mathrm{36.49}$ & $\mathrm{50.25}$ & $\mathrm{\boldsymbol{52.73}}$ & $\mathrm{-}$ & $\mathrm{22.78}$ & $\mathrm{\boldsymbol{36.55}}$ & $\mathrm{-}$ & $\mathrm{-}$ & $\mathrm{-}$ & $\mathrm{-}$\\
    
    % BAM-DETR~\cite{lee2023bam} & 72.93 & \underline{59.95} & 39.38 & 52.33 \\
    % SpikeMba~\cite{li2024spikemba} & 71.24 & \underline{59.65} & 36.12 & 51.74 \\
    % UVCOM~\cite{xiao2024bridging} & - & \underline{59.25} & 36.64 & - \\
    \rowcolor{myblue}
    \textbf{MQVTG (Ours)} & $\mathrm{\boldsymbol{70.97}}$ & $\mathrm{\underline{58.84}}$ & $\mathrm{\boldsymbol{38.84}}$ & $\mathrm{\boldsymbol{51.10}}$ & $\mathrm{49.91}$ & $\mathrm{\boldsymbol{39.09}}$ & $\mathrm{\boldsymbol{25.82}}$ & $\mathrm{\underline{36.15}}$ & $\mathrm{7.18}$ & $\mathrm{\boldsymbol{4.78}}$ & $\mathrm{\boldsymbol{2.35}}$ & $\mathrm{\boldsymbol{5.08}}$ \\
    % \textbf{MQVTG (Ours)} & \textbf{69.92} & 58.20 & 38.66 & \textbf{50.65}  \\
    \toprule
    \end{tabular}
\caption{Moment retrieval results on Charades-STA, TACoS and Ego4D-NLQ.} %with the features from Slowfast and CLIP.}
\vspace{-0.4cm}
\label{table2}
\end{center}
\end{table*}

\begin{table*}[t]\small
 \begin{minipage}[t]{0.4\textwidth}
  \tabcolsep=0.027cm
  \centering
       \begin{tabular}{lcccccc>{\columncolor{myblue}}c}
            \toprule
            \textbf{Method} & Dog & Gym. & Par. & Ska. & Ski. & Sur. & \textbf{Avg.}\\
            % \noalign{\smallskip}
            \hline
            % \noalign{\smallskip}
            \addlinespace[1pt]
            UMT~\cite{liu2022umt} & $\mathrm{65.9}$ & $\mathrm{75.2}$ & $\mathrm{\boldsymbol{81.6}}$ & $\mathrm{71.8}$ & $\mathrm{72.3}$ & $\mathrm{82.7}$ & $\mathrm{74.9}$ \\
            
            QD-DETR~\cite{moon2023query} & $\mathrm{72.2}$ & $\mathrm{\underline{77.4}}$ & $\mathrm{71.0}$ & $\mathrm{72.7}$ & $\mathrm{72.8}$ & $\mathrm{80.6}$ & $\mathrm{74.4}$ \\
            
            UniVTG~\cite{lin2023univtg} & $\mathrm{71.8}$ & $\mathrm{76.5}$ & $\mathrm{73.9}$ & $\mathrm{73.3}$ & $\mathrm{73.2}$ & $\mathrm{82.2}$ & $\mathrm{75.2}$ \\
            
            CG-DETR~\cite{moon2023correlation} & $\mathrm{\underline{76.3}}$ & $\mathrm{76.1}$ & $\mathrm{70.0}$ & $\mathrm{\underline{76.0}}$ & $\mathrm{\underline{75.1}}$ & $\mathrm{81.9}$ & $\mathrm{75.9}$ \\
            
            UVCOM~\cite{xiao2024bridging} & $\mathrm{73.8}$ & $\mathrm{77.1}$ & $\mathrm{75.7}$ & $\mathrm{75.3}$ & $\mathrm{74.0}$ & $\mathrm{82.7}$ & $\mathrm{\underline{76.4}}$ \\
            
            $\text{R}^2$-Tuning~\cite{liu2024r} & $\mathrm{75.6}$ & $\mathrm{73.5}$ & $\mathrm{73.0}$ & $\mathrm{74.6}$ & $\mathrm{74.8}$ & $\mathrm{\underline{84.8}}$ & $\mathrm{76.1}$ \\
            
            % SpikeMba~\cite{li2024spikemba} & $\mathrm{74.4}$ & $\mathrm{75.4}$ & $\mathrm{-}$ & $\mathrm{74.3}$ & $\mathrm{\underline{75.5}}$ & $\mathrm{-}$ & $\mathrm{75.5}$ \\
            
            LLMEPET~\cite{jiang2024prior} & $\mathrm{73.6}$ & $\mathrm{74.2}$ & $\mathrm{72.5}$ & $\mathrm{75.3}$ & $\mathrm{73.4}$ & $\mathrm{82.5}$ & $\mathrm{75.3}$ \\
            
            \rowcolor{myblue}
            \textbf{MQVTG} & $\mathrm{\boldsymbol{76.6}}$ & $\mathrm{\boldsymbol{79.7}}$ & $\mathrm{\underline{75.8}}$  & $\mathrm{\boldsymbol{77.6}}$ & $\mathrm{\boldsymbol{76.4}}$ & $\mathrm{\boldsymbol{84.9}}$ & $\mathrm{\boldsymbol{78.5}}$ \\
            % \textbf{MQVTG (Ours)} & \textbf{69.92} & 58.20 & 38.66 & \textbf{50.65}  \\
            \toprule
        \end{tabular}
        % \captionsetup{margin=0.5cm}
        \caption{Highlight detection results on Youtube HL.}
        \label{table3}%with the features from Slowfast and CLIP.}
  \end{minipage}
  \hspace{0.32cm}
  \begin{minipage}[t]{0.58\textwidth}
   \tabcolsep=0.027cm
   \centering
         \begin{tabular}{lcccccccccc>{\columncolor{myblue}}c}
            \toprule
            \textbf{Method} & VT & VU & GA & MS & PK & PR & FM & BK & BT & DS & \textbf{Avg.}\\
            % \noalign{\smallskip}
            \hline
            % \noalign{\smallskip}
            \addlinespace[1pt]
            TCG~\cite{ye2021temporal} & $\mathrm{85.0}$ & $\mathrm{71.4}$ & $\mathrm{81.9}$ & $\mathrm{78.6}$ & $\mathrm{80.2}$ & $\mathrm{75.5}$ & $\mathrm{71.6}$ & $\mathrm{77.3}$ & $\mathrm{78.6}$ & $\mathrm{68.1}$ & $\mathrm{76.8}$ \\
            
            UMT~\cite{liu2022umt} & $\mathrm{87.5}$ & $\mathrm{81.5}$ & $\mathrm{88.2}$ & $\mathrm{78.8}$ & $\mathrm{81.4}$ & $\mathrm{87.0}$ & $\mathrm{76.0}$ & $\mathrm{86.9}$ & $\mathrm{84.4}$ & $\mathrm{\underline{79.6}}$ & $\mathrm{83.1}$ \\

            QD-DETR~\cite{moon2023query} & $\mathrm{\boldsymbol{88.2}}$ & $\mathrm{87.4}$ & $\mathrm{85.6}$ & $\mathrm{85.0}$ & $\mathrm{85.8}$ & $\mathrm{86.9}$ & $\mathrm{76.4}$ & $\mathrm{91.3}$ & $\mathrm{89.2}$ & $\mathrm{73.7}$ & $\mathrm{85.0}$ \\
            
            UniVTG~\cite{lin2023univtg} & $\mathrm{83.9}$ & $\mathrm{85.1}$ & $\mathrm{89.0}$ & $\mathrm{80.1}$ & $\mathrm{84.6}$ & $\mathrm{81.4}$ & $\mathrm{70.9}$ & $\mathrm{91.7}$ & $\mathrm{73.5}$ & $\mathrm{69.3}$ & $\mathrm{81.0}$ \\
            
            % TR-DETR~\cite{sun2024tr} & 89.3 & 93.0 & 94.3 & 85.1 & 88.0 & 88.6 & 80.4 & 91.3 & 89.5 & 81.6 & 88.1 \\
            CG-DETR~\cite{moon2023correlation} & $\mathrm{86.9}$ & $\mathrm{\underline{88.8}}$ & $\mathrm{\boldsymbol{94.8}}$ & $\mathrm{\boldsymbol{87.7}}$ & $\mathrm{86.7}$ & $\mathrm{\underline{89.6}}$ & $\mathrm{74.8}$ & $\mathrm{\underline{93.3}}$ & $\mathrm{89.2}$ & $\mathrm{75.9}$ & $\mathrm{\underline{86.8}}$ \\
            
            UVCOM~\cite{xiao2024bridging} & $\mathrm{87.6}$ & $\mathrm{\boldsymbol{91.6}}$ & $\mathrm{91.4}$ & $\mathrm{\underline{86.7}}$ & $\mathrm{\underline{86.9}}$ & $\mathrm{86.9}$ & $\mathrm{76.9}$ & $\mathrm{92.3}$ & $\mathrm{87.4}$ & $\mathrm{75.6}$ & $\mathrm{86.3}$ \\
            
            % TaskWeave~\cite{yang2024task} & 88.2 & 90.8 & 93.3 & 87.5 & 87.0 & 82.0 & 80.9 & 92.9 & 89.5 & 81.2 & 87.3 \\
            $\text{R}^2$-Tuning~\cite{liu2024r} & $\mathrm{85.0}$ & $\mathrm{85.9}$ & $\mathrm{91.0}$ & $\mathrm{81.7}$ & $\mathrm{\boldsymbol{88.8}}$ & $\mathrm{87.4}$ & $\mathrm{\underline{78.1}}$ & $\mathrm{89.2}$ & $\mathrm{\underline{90.3}}$ & $\mathrm{74.7}$ & $\mathrm{85.2}$ \\
            
            % LLMEPET~\cite{jiang2024prior} & 90.8 & 91.9 & 94.2 & 88.7 & 85.8 & 90.4 & 78.6 & 93.4 & 88.3 & 78.7 & 88.1 \\
            % SpikeMba~\cite{li2024spikemba} & 85.6 & - & 93.0 & - & 87.1 & 91.5 & - & - & - & - & 85.8 \\
            \rowcolor{myblue}
            \textbf{MQVTG} & $\mathrm{\underline{87.7}}$ & $\mathrm{\boldsymbol{91.6}}$ & $\mathrm{\underline{92.3}}$ & $\mathrm{85.2}$ & $\mathrm{85.7}$ & $\mathrm{\boldsymbol{91.3}}$ & $\mathrm{\boldsymbol{78.5}}$ & $\mathrm{\boldsymbol{96.5}}$ & $\mathrm{\boldsymbol{90.6}}$ & $\mathrm{\boldsymbol{82.9}}$ & $\mathrm{\boldsymbol{88.2}}$ \\
            
            % \textbf{MQVTG (Ours)} & \textbf{69.92} & 58.20 & 38.66 & \textbf{50.65}  \\
            \toprule
        \end{tabular}
        \caption{Highlight detection results on TVSum.}
        \label{table4}%with the features from Slowfast and CLIP.}
   \end{minipage}
   \vspace{-0.4cm}
\end{table*}

\subsection{Training Objectives} \label{3.6}
Our proposed MQVTG operates with losses for moment retrieval and highlight detection. Briefly, we employ L1 and focal objectives for moment retrieval loss $\mathcal{L}_{\text{mr}}$ and use intra-video contrastive loss~\cite{lin2023univtg} as highlight detection loss $\mathcal{L}_{\text{hd}}$. As discussed in Sec.~\ref{3.3}, we use $\mathcal{L}_{\text{mq}} = \mathcal{L}_{\text{cb}} + \lambda_{\text{cmt}}\mathcal{L}_{\text{cmt}}$ as the supervision of moment quantization. We also adopt InfoNCE loss~\cite{oord2018representation} as the alignment loss $\mathcal{L}_{\text{align}}$ to calculate the video-level and layer-wise constraints of temporal encoder $E_t$~\cite{liu2024r}. The overall objectives can be formulated as:
\begin{equation}
    \mathcal{L}_{\text{overall}} = \mathcal{L}_{\text{mr}} + \lambda_{\text{hd}}\mathcal{L}_{\text{hd}} + \lambda_{\text{mq}}\mathcal{L}_{\text{mq}} + \lambda_{\text{align}}\mathcal{L}_{\text{align}},
\end{equation}
where $\lambda_\ast$ are the balancing parameters. Refer to the supplemental material for details about the training objectives.

\section{Experiments}
\label{experiments}
\subsection{Datasets and Metrics}
\label{4.1}
\noindent\textbf{Datasets.}\;
We evaluate the proposed method on six popular video temporal grounding benchmarks, including QVHighlights~\cite{lei2021detecting}, Charades-STA~\cite{gao2017tall}, TACoS~\cite{regneri2013grounding}, Ego4D-NLQ~\cite{grauman2022ego4d}, YouTube Highlights~\cite{sun2014ranking} and TVSum~\cite{song2015tvsum}. Details of each dataset are included in supplemental material.

% QVHighlights spans various themes from everyday lifestyle vlogs to social events in news videos. Charades-STA comprises intricate daily human activities. TACoS mainly showcases long-form videos focusing on culinary activities.

\noindent\textbf{Metrics.}\;
For fair comparisons, we adopt the same metrics with previous works~\cite{moon2023correlation,lin2023univtg}. For QVHighlights, we adopt Recall@1 (R1) with IoU thresholds of 0.5, 0.7, mean average precision (mAP) with IoU thresholds of 0.5, 0.75, and mAP over [0.5:0.05:0.95] for moment retrieval. mAP and HIT@1 where positive samples are defined as with the saliency score of ``Very Good'' are adopted for highlight detection. For Charades-STA, TACoS and Ego4D-NLQ, we use R1 with IoU thresholds of 0.3, 0.5, 0.7, and mean IoU of top-1 predictions. For YouTube Highlights and TVSum, mAP and Top-5 mAP are adopted, respectively.

\subsection{Implementation Details}
\label{4.2}
Following previous methods~\cite{liu2024r,moon2023query}, we employ the pre-trained CLIP model~\cite{radford2021learning} as the spatial encoder and textual encoder. For the encoder-only architecture, we utilize a lightweight recurrent structure~\cite{liu2024r} as the temporal encoder. Without further specification, we employ the encoder-only architecture for the following experiments. We set the embedding dimension $d$ to 256. The size of moment codebook $K$ is set to 1024. All the experiments are implemented on a single Nvidia RTX 3090 GPU. %  consisting of one transformer layer with 8 attention heads,  Please refer to the supplemental material for more details. % Automatic mixed precision (AMP) with FP16 is utilized to accelerate training. \textcolor{red}{Specifically, we extract the last two layers of CLIP encoder features as input to the temporal encoder.}

% Following previous methods~\cite{moon2023query}, for all three datasets, we use SlowFast and CLIP to extract visual features and CLIP to extract textual features. The VGG~\cite{simonyan2014very} and Glove~\cite{pennington2014glove} features are also employed on out-of-distribution splits. %In the encoder, we also use a local regular loss following~\cite{sun2024tr}. We set the embedding dimension $D$ to 256. The number of anchor pairs $\mathcal{K}$ is set to 20 for QVHighlights, 10 for Charades-STA and TACoS. The NMS threshold is set to 0.8. The balancing parameters are set as: $\lambda_{\text{align}} = 0.3$, $\lambda_{\text{iou}} = 1$, and $\lambda_{\text{sal}}$ is set as 1 for QVHighlights, 4 for Charades-STA and TACoS. We train all models with batch size 32 for 200 epochs using the AdamW optimizer with weight decay 1e-4. The learning rate is set to 1e-4. %For more details please refer to supplemental material Sec.~\ref{B2}.%All the experiments are implemented on a single Nvidia RTX 3090 GPU.  and the number of attention heads to 8.

\subsection{Performance Comparison}
\label{4.3}

\paragraph{QVHighlights.} As shown in Tab.~\ref{table1}, we first compare our method to previous methods on QVHighlights, which supports both moment retrieval and highlight detection. Compared to previous methods that focus on learning continuous features, our method achieves the best performance on most metrics. Interestingly, we observe that the performance improvement on highlight detection is less pronounced than on moment retrieval. We attribute this to the difficulty of moment quantization in simultaneously meeting the demands of both moment retrieval (focusing on local relationships) and highlight detection (focusing on global information). % Specifically, for $\text{mAP}_{avg}$ on both splits, MQVTG achieves 48.81\% and 47.08\%, respectively, significantly outperforming the state-of-the-art methods. For a fair comparison, we report results for both the validation and test splits. The notable performance advantages of MQVTG demonstrate the effectiveness of the discriminative information introduced by moment quantization.

\vspace{-0.5cm}
\paragraph{Charades-STA, TACoS \& Ego4D-NLQ.}
We report the results of three moment retrieval datasets in Tab.~\ref{table2}. MQVTG still works better than all previous methods. However, we observe that while our results are notably superior on QVHighlights, the margin is relatively small on these three datasets. We attribute this to low codebook utilization, which is below 10\%—a common issue in previous works~\cite{zheng2023online,takida2022sq}. Additionally, the scene similarity across these datasets requires model to capture fine-grained details. Therefore, codewords intended to provide fine-grained information are not activated, limiting to capture details. We will explore how to improve utilization in the future.
%We attribute this to the similar scenes in the same video of these datasets (in-door and egocentric domains) compared to QVHighlights (various domains), which makes effective clustering harder and discriminative information more difficult to capture.

\begin{table}\small
\tabcolsep=0.1cm
\setlength{\abovecaptionskip}{0.3cm}
\begin{center}
\begin{tabular}{lcccc>{\columncolor{myblue}}c}
\toprule
\multirow{2}{*}{\textbf{Method}} & \multicolumn{2}{c}{R$\mathrm{1}$} & \multicolumn{3}{c}{mAP} \\
\cmidrule(rl){2-3} \cmidrule(rl){4-6}
& @$\mathrm{0.5}$ & @$\mathrm{0.7}$ & @$\mathrm{0.5}$ & @$\mathrm{0.75}$ & Avg. \\ 
% \noalign{\smallskip}
\hline
\addlinespace[2pt]
% $\text{M-DETR}^{\ddagger}$~\cite{lei2021detecting} & $\mathrm{54.19}$ & $\mathrm{34.90}$ & $\mathrm{55.38}$ & $\mathrm{30.38}$ & $\mathrm{32.13}$ \\
% \rowcolor{myblue}
% \hspace{1mm} + \textit{Moment Quantization} & $\mathrm{54.97}$ & $\mathrm{37.16}$ & $\mathrm{57.05}$ & $\mathrm{34.23}$ & $\mathrm{34.59}$ \\
% \hline
% \addlinespace[1pt]
$\text{QD-DETR}^{\ddagger}$~\cite{moon2023query} & $\mathrm{62.58}$ & $\mathrm{46.45}$ & $\mathrm{62.84}$ & $\mathrm{42.44}$ & $\mathrm{41.86}$ \\
\rowcolor{myblue}
\hspace{1mm} + \textit{Moment Quantization} & $\mathrm{63.48}$ & $\mathrm{50.00}$ & $\mathrm{63.28}$ & $\mathrm{44.35}$ & $\mathrm{43.63}$ \\
\hline
\addlinespace[1pt]
$\text{TR-DETR}^{\ddagger}$~\cite{sun2024tr} & $\mathrm{67.16}$ & $\mathrm{51.10}$ & $\mathrm{66.73}$ & $\mathrm{45.37}$ & $\mathrm{44.89}$ \\
\rowcolor{myblue}
\hspace{1mm} + \textit{Moment Quantization} & $\mathrm{67.68}$ & $\mathrm{52.00}$ & $\mathrm{66.97}$ & $\mathrm{46.24}$ & $\mathrm{45.41}$ \\
\hline
\addlinespace[1pt]
$\text{Taskweave}^{\ddagger}$~\cite{yang2024task} & $\mathrm{64.19}$ & $\mathrm{50.77}$ & $\mathrm{64.60}$ & $\mathrm{46.19}$ & $\mathrm{45.30}$ \\
\rowcolor{myblue}
\hspace{1mm} + \textit{Moment Quantization} & $\mathrm{65.16}$ & $\mathrm{51.61}$ & $\mathrm{65.05}$ & $\mathrm{47.21}$ & $\mathrm{45.86}$ \\
\toprule
\end{tabular}
\caption{Generalizability evaluation on QVHighlights \textit{val} split. $\ddagger$ indicates the model is reproduced by the official codebase.}
\label{table5}
\end{center}
\vspace{-0.4cm}
\end{table}

\begin{table}\small
\tabcolsep=0.06cm
\begin{center}
\begin{tabular}{lccc>{\columncolor{myblue}}c}
\toprule
\multirow{2}{*}{\textbf{Method}} & \multicolumn{2}{c}{R$\mathrm{1}$} & \multicolumn{2}{c}{mAP} \\
\cmidrule(rl){2-3} \cmidrule(rl){4-5}
& @$\mathrm{0.5}$ & @$\mathrm{0.7}$ & @$\mathrm{0.5}$ & Avg. \\ 
% \noalign{\smallskip}
\hline
\addlinespace[2pt]
Baseline (w/o quantization) & $\mathrm{65.35}$ & $\mathrm{49.42}$ & $\mathrm{66.99}$ & $\mathrm{45.63}$ \\
% \hspace{1mm} + QBTM (clip quantization) & $\mathrm{66.84}$ & $\mathrm{51.23}$  & $\mathrm{67.32}$ & $\mathrm{47.54}$ \\
\hspace{1mm} + QATM & $\mathrm{66.37}$ & $\mathrm{51.11}$  & $\mathrm{67.43}$ & $\mathrm{47.02}$ \\
\hspace{1mm} + QATM + SQ & $\mathrm{66.52}$ & $\mathrm{51.23}$ & $\mathrm{68.18}$ & $\mathrm{47.54}$ \\
\rowcolor{myblue}
\hspace{1mm} + QATM + SQ + MC (MQVTG) & $\mathrm{\boldsymbol{67.94}}$ & $\mathrm{\boldsymbol{53.03}}$ & $\mathrm{\boldsymbol{68.54}}$ & $\mathrm{\boldsymbol{48.81}}$ \\

% \hspace{1mm} + \textit{codebook calculation} & $\mathrm{67.94}$ & $\mathrm{53.03}$ & $\mathrm{68.54}$ & $\mathrm{51.48}$ & $\mathrm{48.81}$ \\

\toprule
\end{tabular}
\caption{Ablation study on the components of MQVTG. Baseline denotes the model without quantization. `QATM' denotes quantization after temporal modeling. `SQ' denotes soft quantization with continuous features. `MC' denotes moment codebook.}
\label{table6}
\end{center}
\vspace{-0.4cm}
\end{table}

\vspace{-0.5cm}
\paragraph{YouTube Highlights \& TVSum.}
The results of highlight detection on YouTube Highlights and TVSum are reported in Tab.~\ref{table3} and Tab.~\ref{table4}. From an overall perspective, our method gains significant improvements, outperforming the state-of-the-art methods by 2.1\% and 1.4\%, respectively. %achieving an mAP of 78.5\% on YouTube Highlights and 88.2\% on TVSum, Highlight detection can be viewed as a soft version of moment retrieval and could also benefit from the discriminative information of moment quantization.%, as confirmed by experimental results. % In the two datasets, training and inference are domain-specific, which makes it easier for our method to learn discrimination.

\subsection{Generalizability Evaluation}
\label{4.4}
As discussed in Sec.~\ref{3.5}, our moment quantization can be easily integrated into the general encoder-decoder (DETR) architecture. To investigate this property, we conduct experiments by adopting moment quantization on three DETR models: QD-DETR~\cite{moon2023query}, TR-DETR~\cite{sun2024tr} and Taskweave~\cite{yang2024task}. As shown in Tab.~\ref{table5}, our method shows strong adaptability with existing frameworks, demonstrating the effectiveness of moment quantization. %indicated by a consistent performance gain over different baselines,

\subsection{Ablation Study}
\label{4.5}
%To investigate the impact corresponding to key components of the proposed method, we conduct ablation studies on the validation set of QVHighlights.

\paragraph{Main Ablation.}
We first investigate the effectiveness of each component discussed in Sec.~\ref{3.3} and~\ref{3.4}. As shown in Tab.~\ref{table6}, we report the impact according to quantization after temporal modeling, soft quantization, and moment codebook. The results demonstrate that each component contributes significantly to overall performance.% improving by 3.61\% in terms of R1@0.7 and 3.18\% in terms of mAP$_{avg}$. %Next, we conduct detailed ablation experiments in the form of questions.

\begin{table}\small
\tabcolsep=0.16cm
\begin{center}
\begin{tabular}{ccccc>{\columncolor{myblue}}c}
\toprule
\multirow{2}{*}{\textbf{Method}} & \multirow{2}{*}{\textbf{Changes}} & \multicolumn{2}{c}{R$\mathrm{1}$} & \multicolumn{2}{c}{mAP} \\
\cmidrule(rl){3-4} \cmidrule(rl){5-6}
& & @$\mathrm{0.5}$ & @$\mathrm{0.7}$ & @$\mathrm{0.5}$ & Avg. \\ 
% \noalign{\smallskip}
\hline
\addlinespace[2pt]
& Image & $\mathrm{67.16}$ & $\mathrm{51.03}$ & $\mathrm{67.46}$ & $\mathrm{46.55}$ \\
& Clip & $\mathrm{66.84}$ & $\mathrm{51.61}$ & $\mathrm{67.32}$ & $\mathrm{46.93}$ \\
\rowcolor{myblue}
\cellcolor{white} \multirow{-3}{*}{\makecell{Quantization \\ Method}} & Moment & $\mathrm{\boldsymbol{67.94}}$ & $\mathrm{\boldsymbol{53.03}}$ & $\mathrm{\boldsymbol{68.54}}$ & $\mathrm{\boldsymbol{48.81}}$ \\
\hline
\addlinespace[2pt]
& Hard & $\mathrm{67.21}$ & $\mathrm{50.90}$ & $\mathrm{67.85}$ & $\mathrm{47.46}$ \\
& Concat & $\mathrm{67.35}$ & $\mathrm{49.74}$ & $\mathrm{68.20}$ & $\mathrm{47.60}$ \\
& Add & $\mathrm{\boldsymbol{68.39}}$ & $\mathrm{50.84}$ & $\mathrm{68.48}$& $\mathrm{47.88}$ \\
\rowcolor{myblue}
\cellcolor{white} \multirow{-4}{*}{\makecell{D/C Fusion \\ Method}} & Soft & $\mathrm{67.94}$ & $\mathrm{\boldsymbol{53.03}}$ & $\mathrm{\boldsymbol{68.54}}$ & $\mathrm{\boldsymbol{48.81}}$ \\
% \hline
% \addlinespace[2pt]
% & Modulated & $\mathrm{67.81}$ & $\mathrm{52.39}$ & $\mathrm{68.07}$ & $\mathrm{50.29}$ & $\mathrm{48.02}$ \\
% \rowcolor{myblue}
% \cellcolor{white} \multirow{-2}{*}{\textbf{Q3}} & Continuous & $\mathrm{\boldsymbol{67.94}}$ & $\mathrm{\boldsymbol{53.03}}$ & $\mathrm{\boldsymbol{68.54}}$ & $\mathrm{\boldsymbol{51.48}}$ & $\mathrm{\boldsymbol{48.81}}$ \\
\toprule
\end{tabular}
\caption{Ablation study on the quantization method and discrete/continuous features fusion method.}
\label{table7}
\end{center}
\vspace{-0.4cm}
\end{table}

\begin{table}\small
\tabcolsep=0.13cm
\begin{center}
\begin{tabular}{ccccc>{\columncolor{myblue}}c}
\toprule
\multirow{2}{*}{\textbf{Method}} & \multirow{2}{*}{\textbf{Changes}} & \multicolumn{2}{c}{R$\mathrm{1}$} & \multicolumn{2}{c}{mAP} \\
\cmidrule(rl){3-4} \cmidrule(rl){5-6}
& & @$\mathrm{0.5}$ & @$\mathrm{0.7}$ & @$\mathrm{0.5}$ & Avg. \\ 
% \noalign{\smallskip}
\hline
\addlinespace[2pt]
& Random & $\mathrm{67.61}$ & $\mathrm{49.55}$ & $\mathrm{68.91}$ & $\mathrm{46.89}$ \\
& Selection & $\mathrm{\boldsymbol{68.00}}$ & $\mathrm{51.42}$ & $\mathrm{\boldsymbol{69.22}}$ & $\mathrm{47.59}$ \\
\rowcolor{myblue}
\cellcolor{white} \multirow{-3}{*}{\makecell{Codebook \\ Initialization}} & Clutering & $\mathrm{67.94}$ & $\mathrm{\boldsymbol{53.03}}$ & $\mathrm{68.54}$ & $\mathrm{\boldsymbol{48.81}}$ \\
\hline
\addlinespace[2pt]
& Frozen & $\mathrm{66.17}$ & $\mathrm{50.26}$ & $\mathrm{66.62}$ & $\mathrm{46.92}$ \\
& Basic & $\mathrm{67.48}$ & $\mathrm{50.97}$ & $\mathrm{68.33}$ & $\mathrm{47.86}$ \\
\rowcolor{myblue}
\cellcolor{white} \multirow{-3}{*}{\makecell{Codebook \\ Projection}} & Projected & $\mathrm{\boldsymbol{67.94}}$ & $\mathrm{\boldsymbol{53.03}}$ & $\mathrm{\boldsymbol{68.54}}$ & $\mathrm{\boldsymbol{48.81}}$ \\
\hline
\addlinespace[2pt]
& 512 & $\mathrm{\boldsymbol{68.12}}$ & $\mathrm{52.82}$ & $\mathrm{\boldsymbol{68.75}}$ & $\mathrm{48.77}$ \\
\rowcolor{myblue}
\cellcolor{white} \multirow{-1.5}{*}{\makecell{Codebook \\ Size}} & 1024 & $\mathrm{67.94}$ & $\mathrm{\boldsymbol{53.03}}$ & $\mathrm{68.54}$ & $\mathrm{\boldsymbol{48.81}}$ \\
& 2048 & $\mathrm{67.91}$ & $\mathrm{52.89}$ & $\mathrm{68.41}$ & $\mathrm{48.68}$ \\
\hline
\addlinespace[2pt]
& Ch$\rightarrow$QV & $\mathrm{67.54}$ & $\mathrm{52.91}$ & $\mathrm{\boldsymbol{69.19}}$ & $\mathrm{48.45}$ \\
& TA$\rightarrow$QV & $\mathrm{67.72}$ & $\mathrm{52.86}$ & $\mathrm{68.33}$ & $\mathrm{48.63}$ \\
\rowcolor{myblue}
\cellcolor{white} \multirow{-3}{*}{\makecell{Codebook \\ Transferability}} & QV$\rightarrow$QV & $\mathrm{\boldsymbol{67.94}}$ & $\mathrm{\boldsymbol{53.03}}$ & $\mathrm{68.54}$ & $\mathrm{\boldsymbol{48.81}}$ \\
\toprule
\end{tabular}
\caption{Ablation study on moment codebook, including initialization, projection, size and transferability. `QV', `Ch', and `TA' are short for QVHighlights, Charades, and TACoS, respectively.}
\label{table8}
\end{center}
\vspace{-0.4cm}
\end{table}

\begin{figure*}[t]
    \centering
    \setlength{\abovecaptionskip}{0.2cm}
    \includegraphics[height=4.65cm]{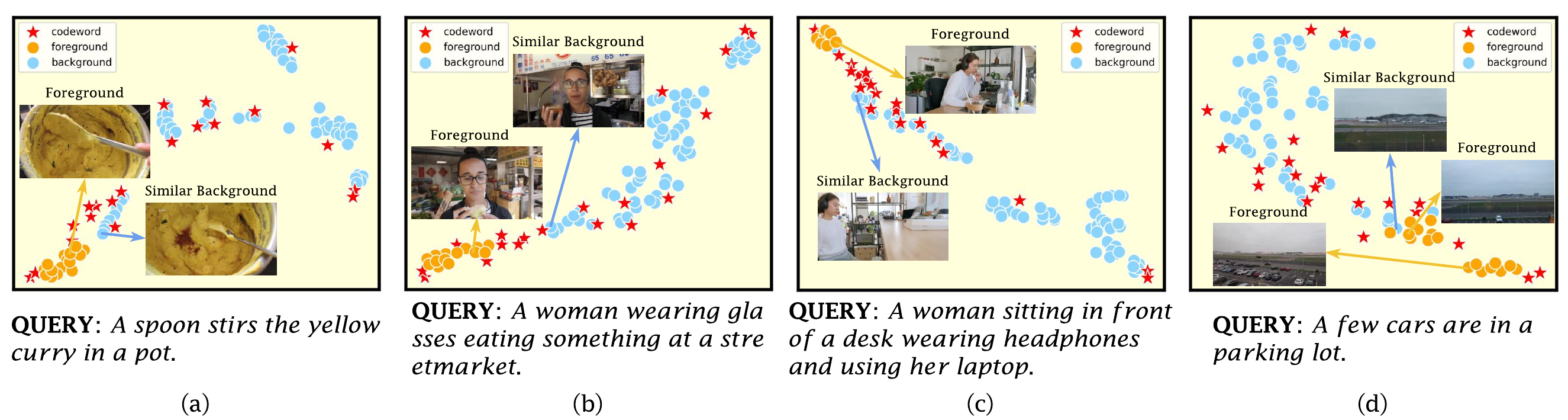}
    \caption{Visulization of effective codebook vectors, foreground and background features in the latent space. (a), (b) and (c) are three successful cases, and (d) is a failure case. For better understanding, we provide both foreground and similar background frames for each example. With codebook assistance, our method performs strong foreground aggregation and fore/background separation across scenarios.}
    \label{fig:4}
    \vspace{-0.2cm}
\end{figure*}

\begin{figure}[t]
    \centering
    \setlength{\abovecaptionskip}{0.3cm}
    \includegraphics[height=2.3cm]
    {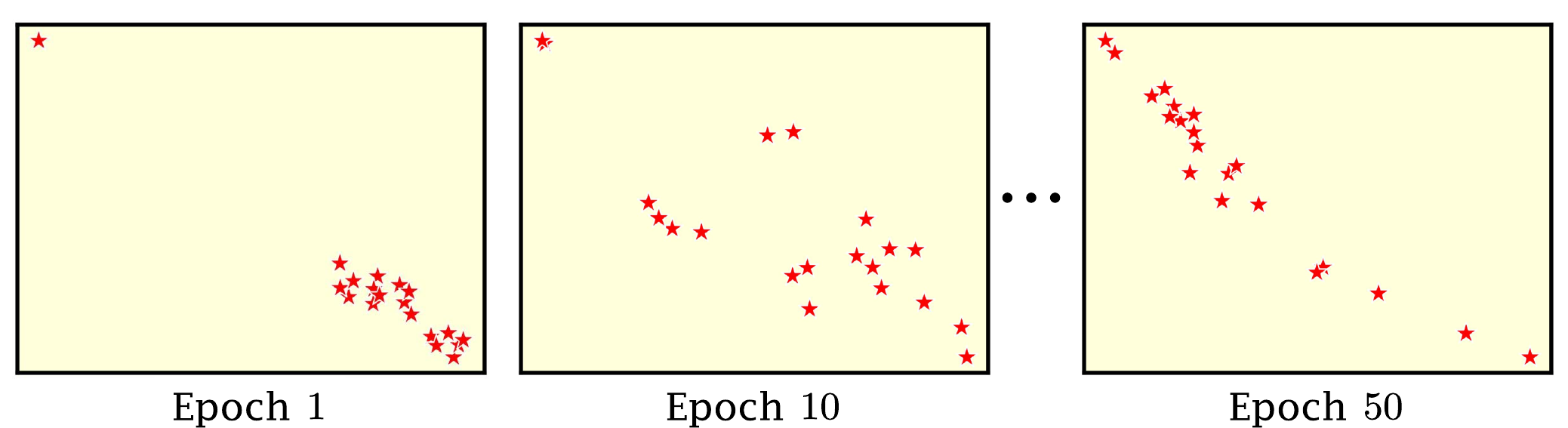}
    \caption{Evolution of effective codebook vectors during training.}
    \label{fig:5}
    \vspace{-0.2cm}
\end{figure}

\vspace{-0.5cm}
\paragraph{Quantization Method.}
In Tab.~\ref{table7}, we explore the effectiveness of three different quantization methods discussed in Fig.~\ref{fig:2}, including image, clip and moment quantization. Image quantization performs patch-level quantization before spatial pooling. Three quantization methods significantly improve performance compared with the baseline results in Tab.~\ref{table6}, indicating that discrete learning from vector quantization benefits the VTG task. Our moment quantization achieves optimal performance through meticulous designs.% The performance gains from clip quantization highlight the need for temporal-level quantization. like quantization after temporal modeling, soft quantization operation, and MQ codebook.

%\noindent\textbf{IoU loss type.}\;
%As shown in Tab.~\ref{table6}, we employ different IoU loss to supervise the IoU score. All loss types can significantly improve performance, among which L2 loss achieves optimal performance.

\vspace{-0.5cm}
\paragraph{Discrete/Continuous Features Fusion Method.}
In Tab.~\ref{table7}, besides hard quantization (directly using discrete features), we design two different fusion modes for continuous and discrete features: ``Add'' means directly adding them and ``Concat" means the concatenation operation. The results show that incorporating discrete features in any manner fails to achieve optimal performance.% likely due to the loss of useful information that affects later localization.

% Clip quantization directly uses discrete features and shows poor performance, while moment quantization employs a soft quantization operation, directly using continuous features. %``Replace'' means directly replacing the continuous feature with the discrete one, as in standard vector quantization;
% \paragraph{Are there other ways to introduce discrete features indirectly?}
% The results indicate that directly using discrete features is ineffective. Therefore, we explore another indirect method in Tab.~\ref{table7}. We follow the modulated quantization operation proposed in~\cite{zheng2022movq}, first normalizing the continuous features, then modulating them with learned scales and biases computed from the discrete features. This modulated quantization is more effective than directly using discrete features but still lags behind our soft quantization.

\vspace{-0.5cm}
\paragraph{Codebook Initialization.}
As shown in Tab.~\ref{table8}, we explore different codebook initialization methods, including random initialization and random selection, where the initial features are randomly chosen from all clip-level features in the dataset. It can be observed that aligning the codebook vectors with continuous video features is crucial. Our $k$-means method generates representative features for initialization, resulting in a better codebook. %initializing with dataset features performs better than random initialization,

\vspace{-0.5cm}
\paragraph{Codebook Projection.}
We provide a comparison between the projected, basic and frozen codebook in Tab.~\ref{table8}, where `frozen' means remaining fixed during training. The results show that the learned codebook, especially incorporating the projector, significantly improves performance.
% Together with clustering initialization, they both create a codebook adaptive for video moments. %During the training of our MQ codebook, we optimize a projector to map the entire codebook to a latent space.

\vspace{-0.5cm}
\paragraph{Codebook Size.}
In Tab.~\ref{table8}, we explore the impact of codebook size. In previous image quantization methods, the size is typically set to 1024, so we adopt this setting. The results show that our method is not sensitive to codebook size.

\vspace{-0.5cm}
\paragraph{Codebook Transferability.}
In Tab.~\ref{table8}, we examine the transferability of our codebook by initializing it with one dataset and training on another. The results indicate that our codebook is not dependent on the specific dataset distribution, demonstrating the robustness of our codebook. %prior initialization. %In our method, both codebook initialization and training are conducted using the same dataset. 

\subsection{Codebook Analysis}
\label{4.6}
\paragraph{High-level Clustering.}
In Fig.~\ref{fig:4}, we provide four examples of the distribution of effective codebook vectors and video features. It can be observed that moment quantization performs higher-level clustering with fewer codebook vectors, directly grouping foreground and background features instead of specific actions or scenes, likely due to lower codebook utilization. As a result, most moments share same codebook vectors. The three successful cases (a), (b), and (c) demonstrate our method's strong foreground aggregation and fore/background separation across different scenarios. In the failure case (d), due to low codebook utilization, it performs poorly in clustering foreground and background that require fine-grained information for discrimination.% We can observe that all features are clustered around the codebook vectors, with foreground and background features showing clear discrimination as they are distributed around different codebook vectors. In the right diagram, even with multiple non-contiguous foreground moments, our method still groups relevant features and separates irrelevant ones.

\vspace{-0.5cm}
\paragraph{Necessity of Soft Quantization.} In Fig.~\ref{fig:4}, our method uses same limited codewords for moments with different semantics. It means that with hard quantization, all moments from different videos would be represented by same codewords, clearly leading to the loss of distinctive information.

\vspace{-0.5cm}
\paragraph{Evolution of Codebook.}
In Fig.~\ref{fig:5}, we visualize the evolution of the distribution of effective codewords during training. As expected, the codewords gradually disperse, allowing them to represent different moments.

\begin{figure}[t]
    \centering
    \setlength{\abovecaptionskip}{0.2cm}
    \includegraphics[height=3.2cm]
    {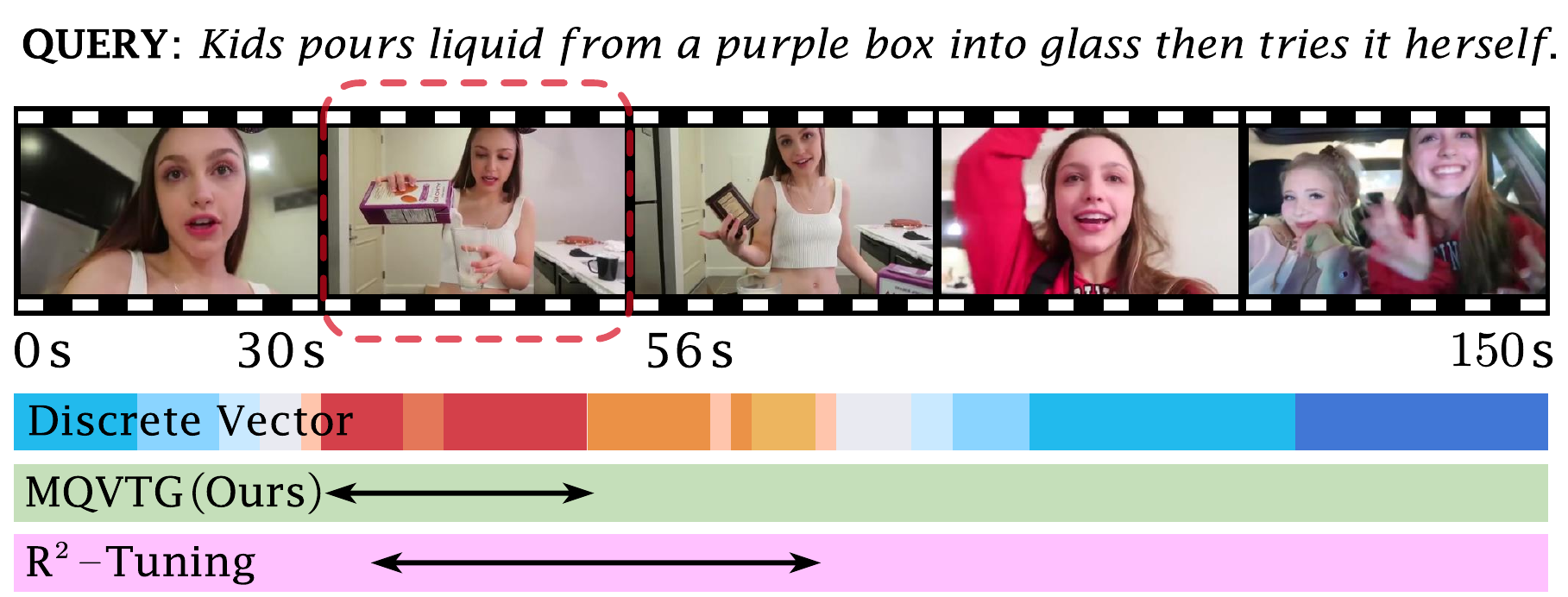}
    \caption{Qualitative result on the QVHighlights. $\text{R}^2$-Tuning is the previous state-of-the-art method.}
    \label{fig:6}
    \vspace{-0.2cm}
\end{figure}

\subsection{Qualitative Result}
\label{4.7}
As shown in Fig.~\ref{fig:6}, we visualize a qualitative result of MQVTG on the QVHighlights dataset. The discrimination introduced by moment quantization allows our method to localize timestamps of the moment precisely. Comparatively, without the discriminative information, $\text{R}^2$-Tuning~\cite{liu2024r} struggles to handle moments in similar scenes.
\section{Conclusion}
In this paper, we propose a Moment-Quantization based Video Temporal Grounding method (MQVTG) to enhance the discrimination between relevant and irrelevant moments. To adapt vector quantization from images to videos, we introduce two progressive implementations, clip quantization and moment quantization, both capable of quantizing video features to capture discriminative information. Clip quantization simply aligns with image quantization, while moment quantization makes significant improvements to meet the cross-clip nature and visual diversity of video moments. Extensive experiments and analysis demonstrate our method effectively groups relevant features and separates irrelevant ones, achieving state-of-the-art performances.

{
    \small
    \bibliographystyle{ieeenat_fullname}
    \bibliography{main}
}

\end{document}